\documentclass{article}
\usepackage[authoryear, sort&compress, round]{natbib}
\usepackage{wrapfig}
\usepackage[dblblindworkshop, final]{neurips_2025}
\workshoptitle{AI for Non-Human Animal Communication}

\usepackage[utf8]{inputenc} % allow utf-8 input
\usepackage[T1]{fontenc}    % use 8-bit T1 fonts
\usepackage{hyperref}       % hyperlinks
\usepackage{url}            % simple URL typesetting
\usepackage{booktabs}       % professional-quality tables
\usepackage{amsfonts}       % blackboard math symbols
\usepackage{nicefrac}       % compact symbols for 1/2, etc.
\usepackage{microtype}      % microtypography
\usepackage{xcolor}         % colors
\usepackage{tablefootnote}
\usepackage{subcaption}
\usepackage{graphicx}
\usepackage{threeparttable}
\usepackage{array}
\usepackage{makecell}
\usepackage{multirow}
\usepackage{tabularx}
\newcommand{\perchvone}{Perch~1.0}
\newcommand{\perchvtwo}{Perch~2.0}

\title{\perchvtwo~transfers `whale' to underwater tasks}
\author{% Author order TBD
Andrea Burns$^*$ \\
Google DeepMind \\
\And
Lauren Harrell\thanks{Corresponding authors: laurenharrell@google.com, andreaburns@google.com}\\
Google Research \\
\And
Bart van Merriënboer \\
Google DeepMind \\
\And
Vincent Dumoulin \\
Google DeepMind \\
\And
Jenny Hamer\\ 
Google DeepMind \\
\And
Tom Denton\\
Google DeepMind 
}

\begin{document}

\maketitle

\begin{abstract}
  \perchvtwo~is a supervised bioacoustics foundation model pretrained on 14,597 species, including birds, mammals, amphibians, and insects, and has state-of-the-art performance on multiple benchmarks. Given that \perchvtwo~includes almost no marine mammal audio or classes in the training data, we evaluate \perchvtwo~performance on marine mammal and underwater audio tasks through few-shot transfer learning. We perform linear probing with the embeddings generated from this foundation model and compare performance to other pretrained bioacoustics models. In particular, we compare \perchvtwo~with previous multispecies whale, \perchvone, SurfPerch, AVES-bio, BirdAVES, and Birdnet V2.3 models, which have open-source tools for transfer-learning and agile modeling. We show that the embeddings from the \perchvtwo~model have consistently high performance for few-shot transfer learning, generally outperforming alternative embedding models on the majority of tasks, and thus is recommended when developing new linear classifiers for marine mammal classification with few labeled examples. 
\end{abstract}

\section{Background: why marine bioacoustics is challenging}
Marine bioacoustics presents different challenges in data collection, verification, and audio properties versus terrestrial sound understanding. Underwater audio data collection requires both specialized sensors (ceramic pressure sensors) and specialized equipment to properly water-seal the electronics with minimal impact on acoustic quality. Deployment requires remote mooring buoys or scuba divers, which are also more expensive than most terrestrial applications. Low visibility under water makes visual confirmation of the vocalizing species impossible in many cases. Finally, the acoustic signal properties also vary depending on depth, temperature, salinity, and bathymetry (underwater topography), where, depending on these conditions and the energy of the signal, sound propagation underwater can be detected over much longer ranges \citep{mercado1999environmental, bass2003physical}. In this context, few-shot learning and transfer learning are valuable tools that allow for quick iteration as new sounds are discovered while not having large amounts of human labeled data. 

Despite these challenges, passive acoustic monitoring is a critical tool for marine conservation and ecology \citep{fleishman2023ecological}, and discoveries about underwater sounds continue to be made. E.g., many mystery sounds discovered in underwater acoustic recordings have been attributed to baleen whale species such as minkes and Bryde's whales years or decades later \citep{rankin2005source, allen2024brydes}. Humpback songs are constantly evolving and move across populations \citep{allen2022song, mercado2012understanding}, but can also be used to distinguish between subpopulations \citep{garland2015population}. Whistle, call, and echolocations are used for determining differences in killer whale specific sub-populations called ecotypes \citep{riesch2011whistle, foote2008variation}. Thus, new machine learning models are needed to classify and annotate large recording databases at scale for underwater sounds as new discoveries are made and songs emerge across populations. 

Recent advances in terrestrial bioacoustics suggest a way forward for rapidly training new models. Previous work has shown excellent few-shot transfer learning---even to new taxonomic groups---using embeddings from bird classifiers~\citep{Ghani2023-mb}. Recently, {\it agile modeling} has been proposed for rapid development of custom bioacoustic classifiers, allowing practitioners to investigate novel fine-grained questions easily~\citep{Dumoulin2025-fj}. First, passive acoustic data is embedded in a vector database. Vector search quickly finds important training examples, and linear probes on pre-computed embeddings are fast to train and evaluate, allowing for an efficient active learning loop.

However, the ability of the embedding model to represent target classes is key to the success of the process. In this paper, we demonstrate that new broadly trained terrestrial bioacoustic models produce high-quality classifiers on marine mammal data. Our goal is to evaluate potential embedding models for marine mammal tasks to provide guidance for developing new classifiers.

\section{\perchvtwo}

Perch is a strong pretrained bioacoustics foundation model~\citep{vanmerrienboer2025perch20bitternlesson} that was trained on over 1.5 million labeled recordings from Xeno-Canto, iNaturalist, the Tierstimmenarchiv, and FSD50K, covering over 14,500 species (mostly birds, but also includes insects, mammals and amphibians). The model uses a convolutional architecture (EfficientNet-B3) trained on log-mel spectrograms using a classification loss. Additionally, the model used a form of self-distillation and a self-supervised loss (in the form of source recording prediction) with the goal of producing strong embeddings that are linearly separable for a wide range of bioacoustics tasks.
Embeddings from the Perch model have shown successful generalization to tasks other than species classification (e.g., individual identification and call type/dialect classification) and to species that were not represented in the training data (e.g., bats).

\section{Datasets}

\perchvtwo~was tested on the Watkins Marine Mammal Sound Database (WMMSD) as part of the BEANS benchmark~\citep{Hagiwara2023-xp}. However, the Watkins database consists of focal recordings from a broad range of conditions (including microphone recordings from ships or land), and thus may not reflect model transfer to passive hydrophone recordings. To establish \perchvtwo{}'s performance for underwater audio tasks more thoroughly for ecology and passive acoustic monitoring use cases, we evaluate its performance on three marine audio validation sets: NOAA PIPAN, ReefSet, and DCLDE 2026.

The DCLDE 2026 dataset \citep{Palmer2025-qk} consists of recordings from multiple providers and hydrophone systems for training models to distinguish between killer whale (\textit{Orcinius orca}) sub-populations and other noises in the northeastern Pacific. The data include over 200,000 annotations averaging 0.73s in length, where 99\% of the annotations are shorter than 3.0s. For evaluation in this work, the annotations are broken into three different label sets as distinct tasks:
\begin{itemize}
    \item Species: humpback, orca, abiotic, and undetermined biological sounds. The specific ecotype annotations for orcas are combined into a single label for this task.
    \item Ecotype: orca-NRKW, -OKW, -SAR, -SRKW, and -TKW. Ecotypes are locally adapted populations with distinct traits, e.g., NRKW refers to northern resident killer whales (orcas) which live in the northeast part of the Pacific Ocean. This classification is only over orcas, as there are no ecotype annotations for humpbacks.
    %Given the subtle nuances in vocalizations across distinct ecotypes, pretrained embeddings that can provide linear separability for distinguishing ecotypes are critically important for orca population monitoring.
    \item Known species: humpback and orca. The specific ecotype annotations for orcas are combined into a single label for this task, and labels for killer whales where the annotator is uncertain are dropped.
\end{itemize}
NOAA PIPAN data is an annotated subset of the NOAA Passive Acoustic Archive~\citep{NOAA-Pacific-Islands-Fisheries-Science-Center2021-jy}. Specifically, it contains recordings from the NOAA Pacific Islands Fisheries Science Center deployments. The labeled segments were extracted from annotations provided in ~\citet{Allen2021-ew, allen2024brydes} in addition to additional annotations provided by NOAA that includes new classes unseen by the models. The classes used in the NOAA PIPAN evaluation set include anthropomorphic noise, unknown whale species, and the following baleen whale species: common minke whale, humpback whale, sei whale, blue whale, fin whale, and Bryde's whale. All examples from NOAA PIPAN are 30s in length and are considered weakly labeled. The ReefSet data are described in detail in~\citet{Williams2025-an}, but include a mix of biological reef noises (such as croaks, crackles, growls), classes for some species/genera (e.g., damselfish, dolphins, and groupers), as well as anthropomorphic noise and wave classes. ReefSet examples are all 1.88s in length. 

\section{Method}

\textbf{Evaluation method.} We use a transfer learning protocol with linear probing to compare embedding performance for each dataset task. This evaluation protocol is established for data that are potentially `weakly' labeled such that the target class is contained in a longer file; for example, the NOAA PIPAN data contains 30-second files labeled at the file level, but we don't know the specific timestamp where the target species vocalization occurs. We apply the same protocol for each model. First, we compute an embedding for each recording in the dataset by chunking it into fixed-sized windows (either 5s or 3s, depending on the model specifications) with the hop size set to the same size as the window, embedding all resulting windows with the model.

Where the labeled recording duration exceeds the window size, we average the embeddings across the windows into a recording-level embedding that is then normalized. Mean pooling the embeddings only applies to NOAA PIPAN for all models, and DCLDE examples longer than 3.0 or 5.0 seconds. We randomly select $k$ recording-level embeddings per class, where $k \in \{4, 8, 16, 32\}$, to train a logistic regression classifier and hold out the remaining embeddings to compute a one-vs-all ROC-AUC metric. We repeat the process 5 times with independently-sampled training sets and report the average ROC-AUC. Thus, we evaluate the embeddings of these models in a few-shot classification setup and do not use the predictions of the models directly. In the development of \perchvtwo{}, these tasks were used as one of the validation sources for model selection \citep{vanmerrienboer2025perch20bitternlesson}.

\begin{table}[h]
\centering
\small
\caption{Embedding model properties}
\label{tab:modelproperties}
\begin{tabular}{ l | c | c | c | c | l } 
\hline
Model name & \thead{Sample\\Rate} & \thead{Window \\Size (s)} & \thead{Embedding\\Dimension} & \thead{\# Model\\Parameters} & Training Taxa  \\
\hline
\perchvtwo{} & 32 kHz & 5.0 & 1536 & 101.8M & Broad Terrestrial \\
\perchvone{} & 32 kHz & 5.0 & 1280 & 23.9M & Birds only \\
SurfPerch & 32 kHz & 5.0 & 1280 & 24.2M & Birds + Reefs \\
GMWM & 24 kHz & 3.0 &  1280 & 4.1M & Whales \\
Birdnet V2.3 & 48 kHz & 3.0 & 1024 & 20.0M & Birds, Frogs \\
AVES-bio & 16 kHz & Variable & 768\footnotemark & 94.4M & General Audio \\
BirdAVES (large) & 16 kHz & Variable & 1024\footnotemark & 315.4M & General Audio + Birds\footnotemark \\
\hline
\end{tabular}
\end{table}
\footnotetext[2]{The AVES-bio output embeddings are computed over a sliding window, providing a 2D array of size ((window size (s) * 49) - 1, 768). We applied mean pooling across the outputs to obtain a single 768 dimensional embedding to be used in linear probing, consistent with the recommendations in the paper.}
\footnotetext[3]{Similarly, mean pooling was applied to the BirdAves data to obtain a single 1024 dimension representation.}
\footnotetext[4]{(Xeno-Canto)}
% \footnote{Embedding is 2D with shape (window size (s)*49 - 1, 768)}.

\textbf{Comparison models.} As our goal is to provide guidance on which pretrained embedding models should be used for agile modeling and transfer learning (with existing tools), we limit our comparisons to models supported in the Perch Hoplite Github repository\footnote{\url{https://github.com/google-research/hoplite}}. We compare the performance of the mean embeddings generated by \perchvtwo{}, \perchvone{}, SurfPerch~\citep{Williams2025-an}, the Google multispecies whale model (GMWM)~\citep{Harvey2024-bh, allen2024brydes}, BirdNet 2.3~\citep{Kahl2021-va}, and the AVES-bio and BirdAVES models~\citep{Hagiwara2023-xx}. A summary of comparison model properties can be found in Table \ref{tab:modelproperties}.

Also note that the published SurfPerch model was trained on the ReefSet data, and similarly, a large portion of the NOAA PIPAN labeled audio data was used to train the GMWM (but the multispecies whale model does not include classes or label sets for sei whales, anthropomorphic noise, or unknown whale). In the NOAA dataset, two classes have fewer examples than can be supported with higher $k$ values, and thus we drop classes for which we have less than $k + 1$ recording-level embeddings. Class `Bm' for $k = 16$ and classes `Bm' and `Be' for $k = 32$ are dropped for the NOAA dataset.

The data for DCLDE 2026 is unseen for all models, although GMWM was trained on killer whales. The scores in~\autoref{tab:underwatertransfer} may slightly differ from previously reported results in the literature due to differences in the linear probe estimation implementation\footnote{In particular,~\citet{Williams2025-an} evaluated the SurfPerch model on ReefSet by training a linear layer using mini-batches and the Adam optimizer. In our experiments we use scikit-learn's \texttt{LogisticRegression} class, which uses the L-BFGS optimizer and weight decay by default.}. \perchvtwo{}'s training data includes around a dozen cetacean recordings from iNaturalist, but these were mostly phone recordings made above water and are not reflective of underwater hydrophone recordings. We bold the highest performance of the non-contaminated models (i.e., those which have no overlapping train/test data).

\begin{table*}[ht]
\centering
\setlength{\tabcolsep}{3.7pt}
\begin{threeparttable}
\small
\caption{Marine learning transfer tasks}
\label{tab:underwatertransfer}
\begin{tabular}{ l | c c c c c c | c c | c c } 
\toprule
% \hline
\multirow{3}{*}{Model}	& \multicolumn{6}{c|}{DCLDE 2026}	&	\multicolumn{2}{c|}{\multirow{2}{*}{NOAA PIPAN}}	&	\multicolumn{2}{c}{\multirow{2}{*}{Reefset}} \\
& \multicolumn{2}{c|}{Species} & \multicolumn{2}{c|}{Ecotype} & \multicolumn{2}{c|}{Known Bio Species} &  & \\
	& $k=8$ & $k=16$ & $k=8$ & $k=16$ & $k=8$ & $k=16$ & $k=8$ & $k=16$ & $k=8$ & $k=16$ \\
\midrule
GMWM & 0.890 & 0.914 & 0.764 & 0.821 & 0.936 & 0.954 & 0.868* & 0.917* & 0.823 & 0.855 \\
SurfPerch &	0.932 & 0.947 & 0.859 &	0.903 & 0.981 & 0.984 & 0.796 &	0.899 & 0.982* & 0.986* \\
\perchvone{} & 0.958 & 0.968 & 0.901 & 0.931 & 0.977 & 0.981 & 0.836 & 0.905 & 0.958 & 0.970 \\
\perchvtwo{} & \textbf{0.970} & \textbf{0.977} & \textbf{0.917} & \textbf{0.945} & 0.983 & 0.989 & \textbf{0.863} & \textbf{0.924} & \textbf{0.975} & \textbf{0.981}  \\
Birdnet V2.3 & 0.942 & 0.959 & 0.905 & 0.933 & \textbf{0.990} & \textbf{0.991} & 0.855 & \textbf{0.924} & 0.965 & 0.974 \\
AVES & 0.880 & 0.916 & 0.825 & 0.879 & 0.965 & 0.971 & 0.825 & 0.893 & 0.972 & 0.979 \\
AVES-Bird & 0.886 & 0.924 &  0.809 & 0.865 & 0.962 & 0.979 & 0.818 & 0.880 & 0.970  & 0.978 \\
\bottomrule 
\end{tabular}
\begin{tablenotes}[para, flushleft]
    \small AUC-ROC of few shot transfer learning comparing \perchvtwo{} to other pretrained models for $k\in\{8,16\}$ samples per class during training. (*) indicates the dataset was in the training data for the embedding model. The Google Multispecies Whale Model (GMWM) was also evaluated in the off-the-shelf classification setting and only reached 0.612 performance. 
    \end{tablenotes}
\end{threeparttable}
\end{table*}

\begin{figure}
\centering
        \includegraphics[width=\textwidth]{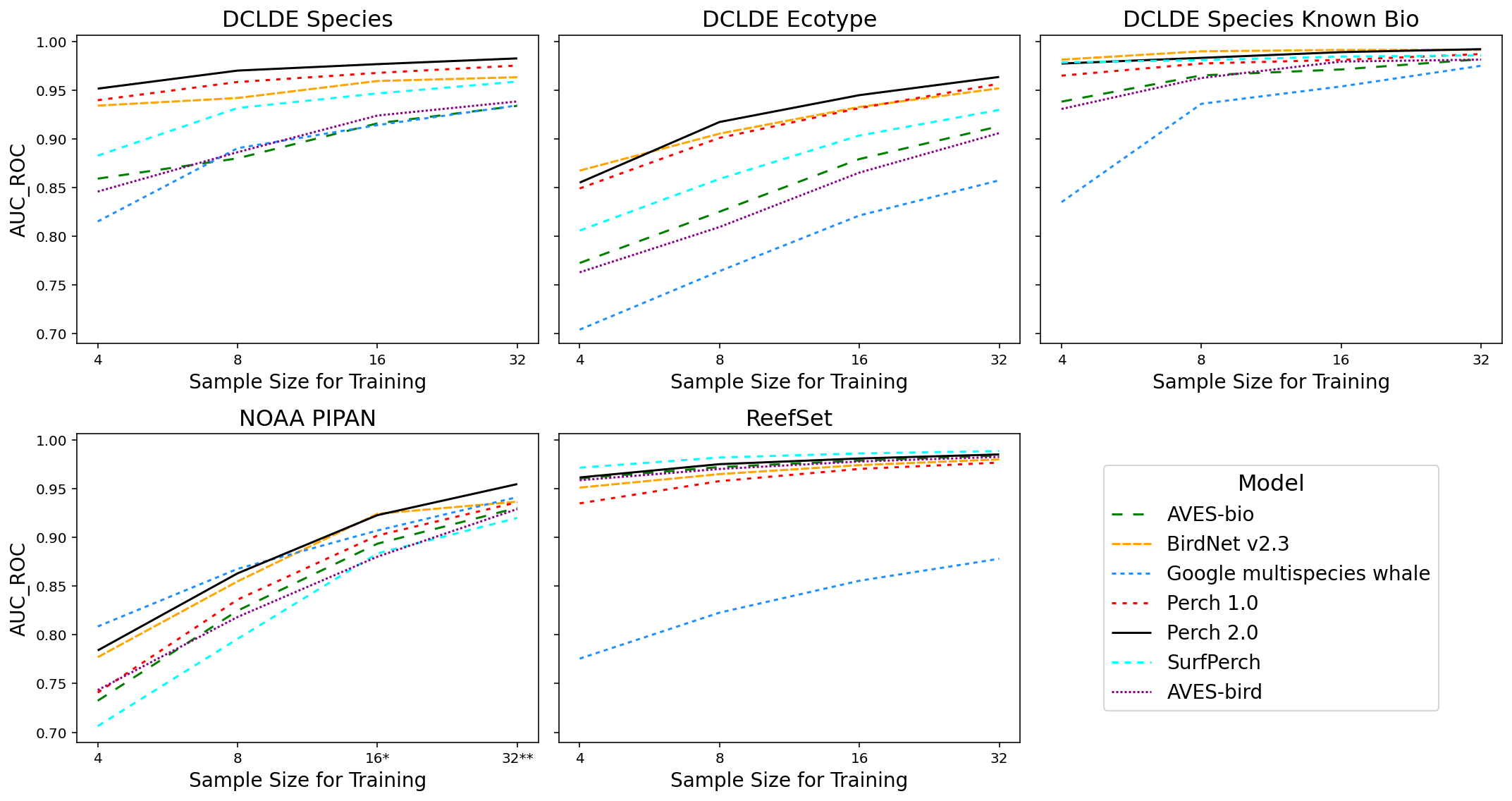} 
\caption{\small Performance of trained models on marine datasets, varying the number of training examples per-class. *Class ``Bm'' dropped for $k=16$; **Classes `Bm' and `Be' dropped for $k=32$ in NOAA PIPAN data.}
\label{fig:auc_roc_by_train_size}
\end{figure}

\section{Results}
Table \ref{tab:underwatertransfer} contains summaries of the results for $k=8$ and $k=16$ and Figure \ref{fig:auc_roc_by_train_size} shows results for $k \in \{4,8,16,32\}$. Two of the selected evaluation datasets were in the training data for two of the models: GMWM was trained on NOAA PIPAN and SurfPerch was trained with ReefSet. We observe that while those models did well on their training datasets, they had worse transfer to new tasks.
Aside from the cases of data contamination, \perchvtwo{} outperforms the published comparison models on ReefSet and all cetacean species tasks (DCLDE 2026 species, ecotypes, and NOAA PIPAN whales), except DCLDE 2026 Known Bio Species, where it comes in second to BirdNet V2.3.

For the DCLDE data, using GMWM's pretrained classification scores for the supported classes yields an AUC-ROC of 0.612. Using the embeddings from this model for few-shot learning the performance jumps to 0.954. The poor performance of the off-the-shelf classification head may be due to overfitting to a particular microphone or other characteristics of the training data.

\section{Discussion - why do models trained on birds work so whale?}
\begin{wrapfigure}{r}{0.5\textwidth}
\centering
        \includegraphics[width=0.45\textwidth]{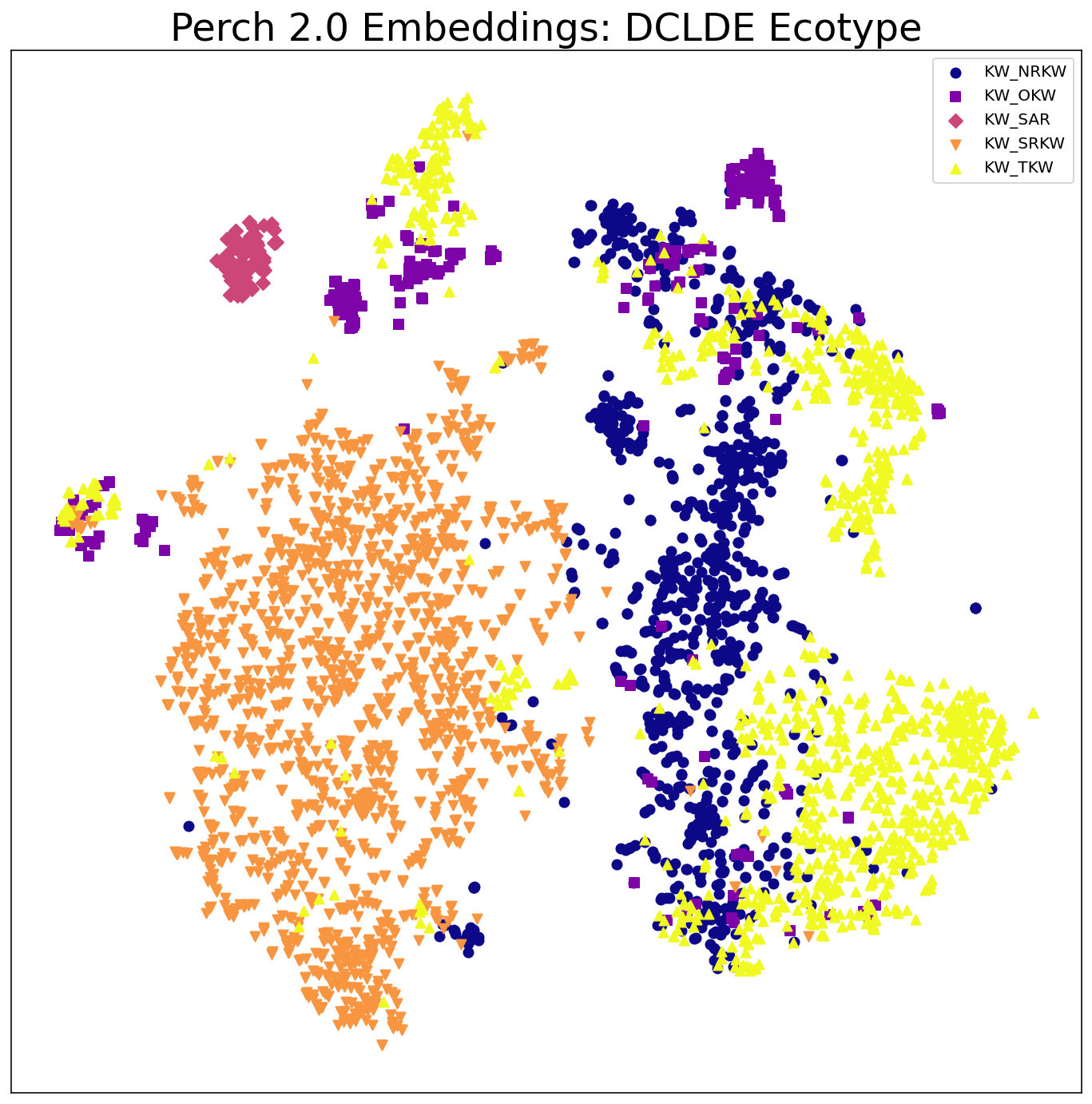} 
\caption{\small tSNE plot of Perch 2.0 Embeddings for DCLDE 2026 Ecotype data}
\label{fig:tsne}
\end{wrapfigure}

The performance of \perchvtwo~and BirdNet on marine mammals is perhaps surprising given the lack of underwater audio included in the training set, but we may hazard an explanation. First, neural scaling laws~\citep{Kaplan2020-fz} tell us that larger models with more training data will most likely perform better. Similarly to how ImageNet-pretrained models dominated computer vision for a long time, it is reasonable to assume that bioacoustics models pretrained on large datasets with thousands of classes can be high performing, even on out-of-domain downstream tasks. Next, the `bittern lesson' learned when training \perchvtwo{} was that bird species classification in particular is a challenging supervision task \citep{vanmerrienboer2025perch20bitternlesson}; the small inter-class variances and large number of classes are likely to force the network to learn detailed acoustic features which are useful for a wide variety of downstream bioacoustics tasks~\citep{Hong2024-ah}. It is also notable that a wide variety of species---including birds and marine mammals---have evolved similar means of sound production, namely the myoelastic-aerodynamic mechanism~\citep{Elemans2015-xy}. This could partially explain why methods transfer remarkably well across these taxa.

In Figure~\ref{fig:tsne}, we illustrate the quality of Perch 2.0's embeddings with a tSNE visualization which plots samples from the DCLDE 2026 ecotype subset. The pooled embeddings are first compressed to dimension 32 with PCA prior to running tSNE. We see that there is strong separability between different ecotypes of the same species (orca), demonstrating that the representations are indeed discriminative. Additional tSNE plots for the other reported models can be found in the Appendix.

While the focus of our efforts was to understand \perchvtwo{}'s performance on underwater transfer tasks, our results show several evaluated embedding models have high performance for transfer-learning, particularly BirdNet. While AVES performance for tasks with fewer examples was often lower, it's possible the method for pooling the embeddings could be modified for improved performance.

Overall, we observe strong few-shot transfer learning performance for general terrestrial bioacoustic models to marine tasks. We expect that using these pretrained models in an agile modeling context will greatly improve the efficiency of a wide range of marine bioacoustic tasks.

\newpage
\section*{Acknowledgments} 
Special thanks to Ann Allen (NOAA Pacific Islands Fisheries Center) and Megan Wood (Saltwater Inc. in support of NOAA Pacific Islands Fisheries Science Center) for providing additional annotations used in the NOAA PIPAN dataset, Isabelle Simpson (Google DeepMind), and Dan Morris (Google Research). 

\bibliography{main.bib}

\newpage
\section{Appendix}\label{sec:appendix}

\begin{figure}[t]
    \centering % Center the entire figure
    \includegraphics[width=\textwidth]{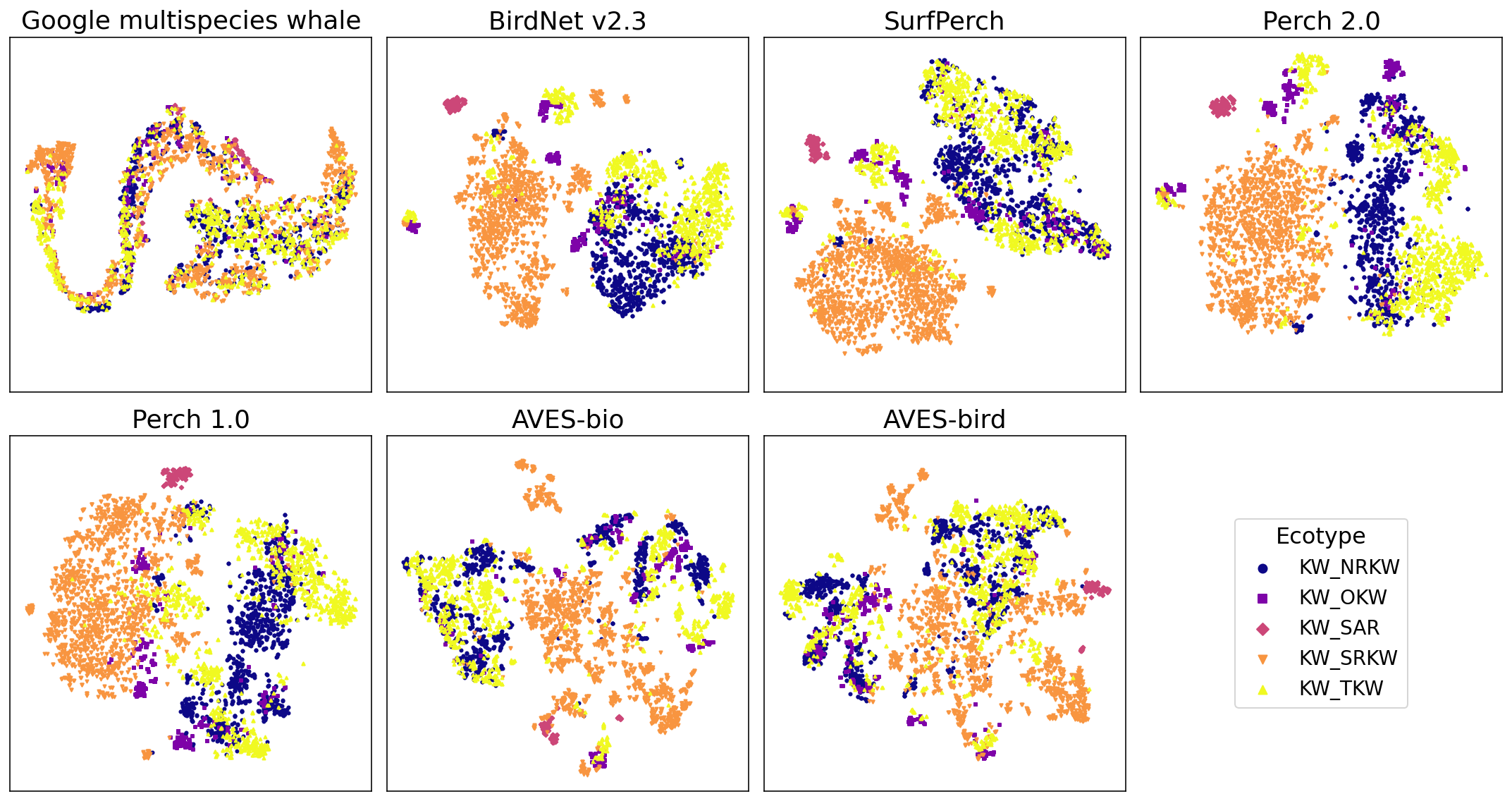}
    % --- The Main Caption for the Entire Figure ---
    \caption{tSNE plots on the DCLDE 2026 Ecotype dataset which contains five ecotype variants of the killer whale (orca) species. Plots were generated with Sklearn PCA and tSNE libraries, with embeddings first projected to 32 dimension vectors prior to tSNE being applied.}
    \label{fig:all_tsne}
\end{figure}

\subsection{Additional tSNE Visualizations}
In Figure~\ref{fig:all_tsne} we include tSNE plots on the DCLDE 2026 Ecotype dataset for all evaluated models (GMWM, SurfPerch, Perch 1.0, Perch 2.0, Birdnet V2.3, AVES-bio, and AVES-Bird). Somewhat unsurprisingly, given the quantitative results of Table~\ref{tab:underwatertransfer}, the Google multispecies whale model embeddings do not show strong linear separability.

All other models show reasonable clustering of orca ecotype, but AVES-bio and BirdAves models show a greater amount of class entanglement, especially with the southern resident killer whale (SRKW) ecotype, which other models have more distinctly separated. Perch 1.0 also does not separate SRKW as well as Perch 2.0, SurfPerch, or BirdNet.

Finally, we see the AVES models do not well separate the transient (TKW) and northern resident killer whales (NRKW). On the other hand, Perch 2.0 appears to have the best boundary between the NRKW and TKW ecotypes.

\end{document}